\documentclass[runningheads]{llncs}
\usepackage{graphicx}
\usepackage{graphics}
\usepackage{xcolor}
\usepackage{amsfonts}
\newcommand*\samethanks[1][\value{footnote}]{\footnotemark[#1]}

\begin{document}
\title{IA-GCN: Interpretable Attention based Graph Convolutional Network for Disease prediction}
%
%
\authorrunning{Kazi et al.}
\author{Anees Kazi\thanks{Equal contribution}\inst{1} \and Soroush Farghadani\samethanks{}\inst{2} \and Nassir Navab\inst{1,3}}

\institute{
Computer Aided Medical Procedures, Technical University of Munich, Germany\\
\and
Sharif University of Technology, Tehran, Iran\\
\and
Whiting School of Engineering, Johns Hopkins University, Baltimore, USA\\
\email{anees.kazi@tum.de}
}
%

%
%
\maketitle
\begin{abstract}


Interpretability in Graph Convolutional Networks (GCNs) has been explored to some extent in computer vision in general, yet, in the medical domain, it requires further examination. Moreover, most of the interpretability approaches for GCNs, especially in the medical domain, focus on interpreting the model in a post hoc fashion. In this paper, we propose an interpretable graph learning-based model which 1) interprets the clinical relevance of the input features towards the task, 2) uses the explanation to improve the model performance and, 3) learns a population level latent graph that may be used to interpret the cohort’s behavior. In a clinical scenario, such a model can assist the clinical experts in better decision-making for diagnosis and treatment planning. The main novelty lies in the interpretable attention module (IAM), which directly operates on multi-modal features. Our IAM learns the attention for each feature based on the unique interpretability-specific losses. We show the application on two publicly available datasets, Tadpole and UKBB, for 3 tasks of disease, age, and gender prediction. Our proposed model show superior performance with respect to compared methods with an increase in an average accuracy of 3.2\% for Tadpole and 1.6\% for UKBB Gender, and 2\% for the UKBB Age prediction task. Further, we show exhaustive validation and clinical interpretation of our results.

\end{abstract}


\section{Introduction}
Recently, Graph Convolutional Networks(GCNs) have shown great impact in the medical domain, especially for multi-modal data analysis \cite{parisot2018disease,gopinath2019graph,kazi2019self}. So far, GCNs have been applied to many medical problems such as Autism and Alzheimer's prediction \cite{anirudh2019bootstrapping}, brain shape analysis \cite{gopinath2019adaptive}, handling missing data \cite{vivar2019multi}, Pulmonary Artery-Vein Separation \cite{zhai2019linking}, functional connectivity analysis \cite{kim2020understanding}, mammogram analysis \cite{du2019zoom} and brain imaging \cite{stankevivciute2020population,gopinath2019graph}.

GCNs allow a setup where a cohort of patients can be represented in terms of a population graph. It is done usually by putting the patients as nodes and neighborhood connections computed based on the similarity between the corresponding medical meta-data \cite{parisot2017spectral}. Multi-modal features for each patient are then initialized at the corresponding node of this population graph. Graph convolutions are then applied to this setup to process the node features for optimizing various objectives.
In recent literature, many methodological advances have been made, especially for medical tasks, such as dealing with the out-of-sample extension \cite{hamilton2017inductive}, handling multiple graph scenario \cite{vivar2020simultaneous,kazi2019self,kazi2019graph}, graph learning \cite{cosmo2020latent}, and dealing with graph heterogeneity \cite{kazi2019inceptiongcn}. However, GCNs are complex models and highly sensitive towards the input graph, input node features, and task (loss) \cite{jin2020certified,wang2019graphdefense}. Interpreting the model's outcome with respect to each of them is essential and yet a challenging task. In spite of their great success, GCNs are still less transparent models. Interpretability techniques dealing with the analysis of GCNs with respect to the above factors are gaining importance only since the last couple of years.
GNNExplainer \cite{ying2019gnnexplainer}, from the computer vision domain, for example, is one of the pioneer works in this direction. This paper proposes a post hoc technique to generate an explanation of the Graph Convolutional(GC) based model with respect to input graph and features. This is obtained by maximizing the mutual information between the pre-trained model output and the output with selected input sub-graph and features. Further conventional gradient-based and decomposition-based techniques have also been applied to GCN in \cite{baldassarre2019explainability}. Another work \cite{huang2020graphlime} proposes a local interpretable model explanation for graphs. It uses a nonlinear feature selection method leveraging the Hilbert-Schmidt Independence Criterion (HSIC). However, the method is computationally complex as it generates a nonlinear interpretable model.

In the medical domain, interpretability is essential as it can help in informed decision-making during diagnosis and treatment planning.
However, works targeting interpretability, especially for GCNs, are limited. One recent work \cite{jaume2020towards} proposes a post hoc approach similar to  the GNN explainer applied to digital pathology. Another method Brainexplainer \cite{li2020braingnn} proposes an ROI-selection pooling layer (R-pool) that highlights ROIs (nodes in the graph) important for the prediction of neurological disorders. Interpretability for GCNs is still an open challenge in the medical domain. 

In this paper, we not only target the interpretability of GCNs but also design a model capable of incorporating the interpretable attentions for better model performance. Unlike conventional GCN models that require a pre-defined graph, our model does not require a pre-computed graph. As such pre-defined graphs may be noisy or unknown. In the recent graph deep learning literature, only a few works have focused on latent graph learning problems \cite{kazi2019self,kazi2019graph,coates2019mgmc}. The proposed model is empowered with an interpretability-specific loss that helps in increasing the confidence of the interpretability. 
We show that such an interpretation-based feature selection enhances the model performance.

\textbf{Contributions:} In this work, we propose a GCN-based method that is capable of 1) learning the Interpretable Attention for input features, 2) using these Interpretable Attentions to better optimize the task better, and 3) providing an explanation for the model's outcome. We leverage a unique Loss function for interpretability.  Further, our model learns and outputs a population-level graph that aids in discovering the relationship between the patients. The whole pipeline is trained end-to-end for a) task at hand (disease, age, and gender prediction), b) learning the latent population-level graph, and 3) interpretable attention for features. 
We show the superiority of the proposed model on 2 datasets for 3 tasks, Tadpole \cite{marinescu2018tadpole} for Alzheimer's disease prediction (three-class classification) and UKBioBank (UKBB) \cite{miller2016multimodal} for gender (2 classes) and age (4 classes) classification task. We show several ablation tests and validations supporting our hypothesis that incorporating the interpretation with the model is beneficial. In the following sections, we first give mathematical details of the proposed method, then describe the experiments, followed by discussion and conclusion.
\section{Method}
Given the dataset $Z=\left [X,Y\right]$, where $X\in \mathbb{R}^{N \times D}$ is the feature matrix with dimension $D$ and $N$ patients. $Y \in \mathbb{R}^{N}$ being the labels with $c 
$ classes. The task is to classify each patient into the respective class from $c$. Along with this primary task, the proposed method targets to obtain two more by-products 1) population-level graph that explores the clinically semantic latent relationship between the patients, 2) Interpretable Attentions for features and incorporation of them in an end-to-end model for better representation learning. Both graph learning and feature interpretation are trained end-to-end on unique loss that contributes towards the three tasks. 
\begin{figure}[t]
	\centering
	\includegraphics[width=0.8\textwidth]{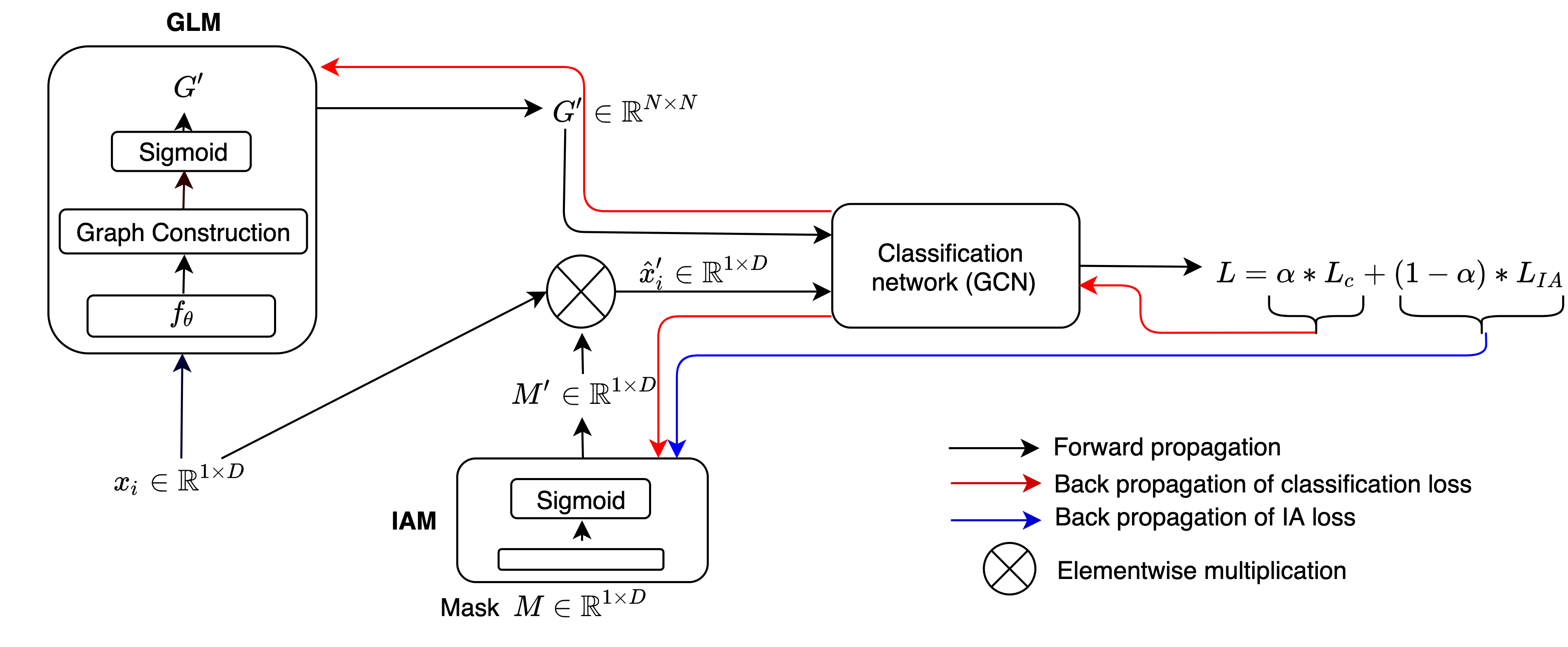}
	\caption{IA-GCN architecture. The model consists of three components: 1) IAM, 2) GLM, and 3) Classification Module. These are trained in an end-to-end training fashion. In backpropagation, two loss functions are playing roles which are demonstrated in blue and red arrows.}
	\label{fig.method}
\end{figure}
In the following paragraphs, we will explain Graph Learning Module (GLM), Interpretable attention Module (IAM), and Classification module.\\  
\textbf{Graph Learning Module (GLM)}

Given the dataset $Z$, the main objective of GLM is to predict an optimal graph/ adjacency matrix  $G'\in \mathbb{R}^{N\times N}$.  $G'$ is then used in the graph convolutional network for the classification, as shown in figure \ref{fig.method}. The major challenge in solving the graph learning task is the discrete nature of the graph edges. This makes the backpropagation impracticable, making the model non-differentiable. Inspired from DGM \cite{cosmo2020latent} we define our GLM in 2 steps as follows:
\textbf{Step 1:} Defining $f_{\theta}$ to predict lower-dimensional embedding from the input features for the edges. $f_{\theta}$  can be defined as $ \hat{X} = f_{\theta}(X, \theta)$
where $f$ is a generic function with learnable parameters $\theta$. To keep it simple, we chose $f$ to be MLP, however, could be changed based on the input data and the task.  $f_{\theta}$ takes the feature matrix $X\in \mathbb{R}^{N\times D}$ as input and produces $\hat{X}$ embedding specific for the graph as output. It should be noted that $f_{\theta}$ is dedicated to learning the feature embedding $\hat{X}$ only to predict the graph structure. \textbf{Step 2:} We compute a fully connected graph with continuous edge values (shown as graph construction in figure \ref{fig.method}) using the euclidean distance metric between the feature embedding  $\hat{x_{i}}$ and $\hat{x_{j}}$ obtained in step 1. 
We use the sigmoid function for soft thresholding keeping the GLM differentiable . $g'_{ij}$ is computed as $g'_{ij}=\frac{1}{1+e^{ \left(t \|\mathbf{\tilde{x}_{i}}-\mathbf{\tilde{x}_{j}}\|_2 + \theta \right)} }$ where $\theta$ being the threshold parameter and $t$ the temperature parameter pushing values of $a_{ij}$ to either $0$ or $1$. Both $t$ and $\theta$ are optimized during training.
The output $G'$ from GLM is further given as an input to the classification module.\\
\textbf{Interpretable Attention Module (IAM)}

For IAM, we define a differentiable and continuous mask $M \in \mathbb{R}^{1 \times D}$ that learns an attention coefficient for each feature element from $D$ features. IAM can be mathematically defined as $x_i' =\sigma (M) \times x_i
$
where, $x_i'\in \mathbb{R}^{1\times D}$ is the masked output, $\sigma$ is the sigmoid function.  
$\sigma (M)$ represents the learned Mask $M' \subset \left [ 0,1 \right ]$.
Our mask $M$ is continuous as the target is to learn the interpretable attentions while the model is training. 
Conceptually, the corresponding weights in the mask $M'$ should take a value close to zero when a particular feature is not significant towards the task. In effect, $m_i$ corresponding to $d_i$ may improve or deteriorate the model performance depending on the importance of $d_i$ towards the task. Further, the proposed IAM is trained by a conventional classification loss $L_{c}$ and two separate regularization losses $F_{MSL}$, and $F_{MEL}$.
%
All the losses are defined in detail in the loss function section.
\\
\textbf{Classification Module with joint optimization of GLM and IAM}

As mentioned before, the primary goal is to classify each patient $x_{i}$ into the respective class $y_{i}$. The classification model can be mathematically defined as $\hat{Y} = f_{\theta }\left ( X', G' \right )$
where $f_{\theta }$ is the classification function with learnable parameters $\theta$, $G'$ being the learned latent population graph structure, $M'$ being the interpretable attention mask. We define $f_{\theta }$ as a generic graph convolution network targeted towards node classification.\\ 
\textbf{Interpretability focused loss functions}

In order to optimize the whole network in an end-to-end fashion, we define the loss function as $L = \alpha*L_{c} + (1- \alpha)* L_{IA}$ 
where $L$ is the final loss function used to optimize the whole model. $L_{c}$ is the classification loss. $L_{IA}$ is interpretable attention loss. Furthermore, $\alpha$ is the weighting factor, chosen empirically. 

We use softmax cross-entropy for classification loss $L_{c}$. Training the model with only $L_c$ has three main drawbacks 1) the model performance is not optimal, 2) the mask learns average value for all the features depicting the uncertainty, and 3) the unimportant features take higher weights in the mask. In order to overcome these drawbacks, we define $L_{IA}$ as $L_{IA} = \alpha_1 * F_{MEL}+ \alpha_2 * F_{MSL}$
, where $F_{MEL}$ is the feature mask entropy loss, and $F_{MSL}$ is the feature mask size loss as defined in section 3.2. $\alpha_1 $, $\alpha_2$ are the weighting factors chosen empirically. Firstly, $F_{MSL}= \sum_{i=0}^{D-1} m'_i$ nails the values of individual $m'_i$'s to minimum. Otherwise, all the features get the highest importance with all the $m'$s taking up the value 1. On the other hand, $ F_{MEL} = \sum_{i=0}^{D-1} -m'_i log (m'_i)$ tries to push values away from $0.5$, which makes the model more confident about the importance of the feature $d_i$.
Rewriting the loss function as:
\begin{equation}
   L = (1- \alpha)L_{c} + \alpha* (\alpha_1 * \sum_{i=0}^{D-1} -m'_i log (m'_i)+ \alpha_2 * \sum_{i=0}^{D-1} m'_i)
\end{equation}
We would like to point out \textbf{the role} of each element in final loss $L$. Each term in $L$ affects each other. The final values of each term are determined by the optimal point of $L$. $F_{MSL}$ tried to bring $m'_i$ to zero. However, the important features must get corresponding $m'=1$, which is possible due the effect of $L_c$ on $F_{MSL}$.  Further, $F_{MEL}$ tried to move the values of $m'$ away from 0.5. However, it is governed by $L_c$ and $F_{MSL}$ together to do so. All other weighting factors $\alpha$, $\alpha_1$ and $\alpha_2$ are chosen empirically for optimal $L$.  
\section{Experiments}
We conduct our experiments on two publicly available datasets for three tasks. Tadpole \cite{marinescu2018tadpole} for Alzheimer's disease prediction and UK Biobank\cite{miller2016multimodal} for age and gender prediction.
The task in \textbf{Tadpole} dataset is to classify 564 patients into three categories (Normal, Mild Cognitive Impairment, Alzheimer's) which represent their clinical status. Each patient has 354 multi-modal features cognitive tests, MRI ROIs, PET imaging, DTI ROI, demographics, and others. 
On the other hand, the \textbf{UK Biobank} dataset consists of 14,503 patients with 440 features per individual, which are extracted from MRI and fMRI images. Two classification tasks are designed for this dataset 1) gender prediction, 2) categorical age prediction.
In the second task, patients’ ages are quantized into four decades as the classification targets.

We compare the proposed method with four state-of-the-art and one baseline method shown in Table 1. We perform an experiment with a linear classifier to see the complexity of the task. Further, Spectral-GCN \cite{parisot2018disease} is the Chebyshev polynomial-based GC method, and Graph Attention Network(GAT)\cite{velivckovic2017graph} is a spatial method. We compare with these two methods as they require a pre-defined graph structure for the classification task. Whereas our method and DGM\cite{kazi2020differentiable} learns a graph during training. In such methods, the graph structure is constructed based on the patients' similarities either from the imaging features or the meta-data.
Our argument is that pre-computed/ preprocessed graphs can be noisy, irrelevant to the task or may not be existing.
Depending on the model, learning the population graph is much more clinically semantic. Unlike Spectral-GCN and GAT, DGCNN\cite{wang2019dynamic}, constructs a Knn graph at each layer dynamically during training. This removes the requirement of a pre-computed graph. 
However, the method still lacks the ability to learn the latent graph.\\
\textbf{Implementation Details:} $M$ is initialized either with Gaussian normal distribution or constant values. Experiments are performed using Google Colab with Tesla T4 GPU with PyTorch 1.6. Number of epochs = 600. Same 10 folds with the train: test split of 90:10 are used in all the experiments. Two MLP layers (16→8) for GLM. Ttwo Conv layers followed by a FC layer (32→16→\# classes) for classification network. RELU is used as the activation function.\\
\textbf{Classification performance}

Three different experiments are performed for the comparison between the methods. We report Accuracy, AUC and F1 score of each experiment. In the first experiment, We compare the methods on Tadpole as shown in Table. 1. The lower F1-score indicates the challenging task due to the class imbalance. ($\frac{2}{7}$,$\frac{4}{7}$,$\frac{1}{7}$)
Our proposed method outperforms the state-of-the-art. The low variance of the proposed method shows the stability of the method.\\ 
\begin{table}[t]
\centering
  \label{tab:tadpole}%
  {\caption{Performance of the proposed method compared with several state-of-the-art methods, and various baselines methods on Tadpole dataset. LC stands for Linear Classifier.}}%
  {\begin{tabular}{|c|ccc|} 
\hline
Method & Accuracy & AUC &F1 \\
\hline
LC \cite{Hoerl1}& 70.22 ± 06.32 & 80.26 ± 04.81 & 68.73 ± 06.70\\ 
GCN\cite{parisot2017spectral} & 81.00 ± 06.40 & 74.70 ± 04.32&78.4 ± 06.77 \\ 
GAT\cite{velivckovic2017graph} & 81.86 ± 05.80 & 91.76 ± 03.71 &80.90 ± 05.80\\
DGCNN\cite{wang2019dynamic}&  84.59 ± 04.33 & 83.56 ± 04.11   &82.87 ± 04.27\\
DGM\cite{cosmo2020latent} & 92.92 ± 02.50& 97.16 ± 01.32&91.4 ± 03.32\\ 
IA-GCN & \textbf{96.08 ± 02.49} & \textbf{98.6 ± 01.93}&\textbf{94.77 ± 04.05} \\ 
\hline
\end{tabular}}
\end{table}
Two more tasks of gender and age prediction are performed with the above methods on UKBB dataset with much larger dataset size. 
The results are shown in Table. \ref{tab:ukbb}. Our method shows superior performance and AUC reconfirms the consistency of model's performance.
For age prediction, results demonstrate that the overall task is much more challenging than the gender prediction. Lower F1-score shows the existence of class imbalance. Our method outperforms the DGM (SOTA) by $2.02\%$ and $1.65\%$ in accuracy for Gender and Age task respectively. 
\begin{table}[t]
\centering
  \label{tab:ukbb}%
  \caption{Performance of the proposed method compared with several state-of-the-art methods, and a baselines methods on UKBB dataset. \textbf{Top:} Gender classification, \textbf{Bottom:} Age classification.}%
\begin{tabular}{|c|c|c|c|c|c|c|} 
\hline
& & LC& GCN&DGCNN&DGM&IA-GCN\\ 
\hline
&Accuracy &81.70±01.64 &83.7±01.06&87.06±02.89&90.67±01.26&92.32±00.89\\ 
Gender&AUC & 90.05±01.11&83.55±00.83&90.05±01.11&96.47±00.66&97.04±00.59\\ 
&F1& 81.62±01.62 &83.63±00.86&86.74±02.82&90.65±01.25& 92.25±00.87\\
\hline
&Accuracy &59.66±01.17  &55.55±01.82&58.35±00.91&63.62±01.23 &65.64±01.12\\ 
Age&AUC & 80.26±00.91&61.00±02.70 &76.82±03.03&76.82±03.03&76.82±03.03\\ 
&F1&48.32±03.35&40.68±02.82&47.12±03.95&50.23±02.52&51.73±02.68\\
\hline
\end{tabular}
\end{table}

The above results indicate that graph-based methods perform better than the conventional ML method (LC). The incorporation of graph convolutions helps in better representation learning resulting in better classification.
Further, GAT requires full data in one batch along with the affinity graph, which causes Out Of Memory in UKBB experiments. Moreover, DGCNN and DGM achieve higher accuracy compared to Spectral-GCN and GAT. This confirms our hypothesis that a pre-computed graph may not be optimal. Between DGCNN and DGM, DGM performs better than DGCNN, confirming that learning graph is beneficial to the final task and for getting latent semantic graph as output (Table. 2).\\ 
\begin{table}[t]
\parbox{.6\linewidth}{
\centering
\label{tab:val alpha}
\begin{tabular}{|c|c|c|c|} 
    \hline
    &ACC&AUC&F1\\
    \hline
    a) &89.2±05.26&96.47±02.47&88.60±05.32\\
    b) &92.92±02.50&97.16±01.32&91.4±03.32\\
    c) &79.70±04.22&90.66±02.64&77.9±6.38\\
    d) &95.09±03.15&98.33±02.07&93.36±03.28\\
    \hline
    \end{tabular}
    \vspace{10px}
\caption{Performance shown for classification task on Tadpole dataset. We show results for four baselines on DGM\cite{cosmo2020latent} with different input feature setting.}
}
\hfill
\parbox{.35\linewidth}{
\centering
\label{tab:validation1}
\begin{tabular}{ |c|c|c|c| } 
\hline
$\alpha$& Accuracy& Avg.4 & Avg.O\\
\hline
0  &  57.00±09.78 &$10^{-6}$& $10^{-6}$\\ 
0.2 & 94.20±03.44 &0.12& 0.002 \\ 
0.4 & 95.10±02.62&0.29& 0.001\\ 
0.6 & 96.10±02.49& 0.74& 0.0\\
0.8 &  95.80±02.31&0.78& 0.23\\
1.0 &   95.40±02.32&0.82& 0.42\\ 
\hline

\end{tabular}
    \vspace{5px}
\caption{Performance of IA-GCN on Tadpole dataset in different settings.}
}
\end{table}
\textbf{Interpretability}

We now check the importance of the selected features by manually adding and removing the features from the input for 
DGM\cite{cosmo2020latent}. We chose the best performing method DGM for this experiment. In Table. 3, we show experiments with different choices on input features. The four variations of input are given as a) method trained traditionally. b) ridge classifier used to reduce the dimensionality at the pre-processing step,  c) method trained conventionally with all input features except the features selected by the proposed method, d) model trained on only features selected by the proposed method. The following can be observed from Table \ref{tab:validation1}. In general, the feature selection technique (b and d) is beneficial for the task. The proposed IA-based feature selection performs the best for the method (d). When the models are trained with features other than the selected one, the performance of the model drastically drops.\\
\textbf{Analysis of the loss function}

Further, we investigated the contribution of two main loss terms $L_c$ and $L_{IA}$ and $F_{MEL}$, $F_{MSL}$, towards the optimization of a task. Table. 4 shows changes in the performance and the average of attention values (Avg.4 and Avg.O) with respect to the changes on $\alpha$. We look into the contribution of each loss term ($L_c$, $L_{IA}$). It can be seen that the performance drops significantly with $\alpha=0$. Best accuracy at $\alpha= 0.6$ shows that both the loss terms are necessary for the optimal performance of the model. We report the average attention for the top four features (Avg.4) selected by the model and other features (Avg.O). 
The Avg.O surges dramatically each time $\alpha$ increases. This proves that $L_{IA}$ plays the most important role in shrinking the attention values of less important features. 

\begin{figure}[t]
	\centering
	\includegraphics[width=0.1\textwidth]{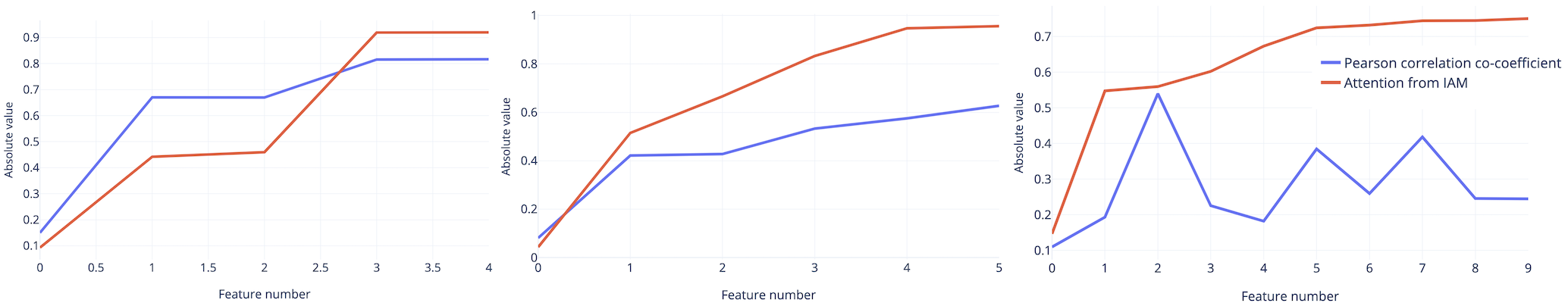}
	\caption{Person Correlation Coefficient with respect to groundtruth vs. attention learned by the IAM. \textbf{(left)}Tadpole, \textbf{(center)} UKBB for gender prediction. \textbf{(right)} UBBB for age prediction. The names of the features are provided in the supplementary material.}
	\label{coeff}
\end{figure}

In Figure. \ref{coeff}, we show the attention learned by IAM vs. the Pearson correlation coefficient for the corresponding features with the ground truth. For both Tadpole and UKBB datasets, it is observed that 1) the set of selected features are different depending on the task, 2) age classification is more difficult than gender classification; thus, more features are given attention.\\
\textbf{Clinical interpretation} 

In the Tadpole, our model selects four cognitive features. 
Cognitive tests measure cognitive decline in a straightforward and quantifiable way. Therefore, these are important in Alzheimer's disease prediction \cite{jack2013biomarker}, in particular CDRBS and MMSE \cite{o2008staging,arevalo2015mini}.
Apart from cognitive tests, Tadpole dataset includes other imaging features. In the absence of cognitive features, the model selects the PET FDG feature.
The PET measures are one of the earliest biomarkers of Alzheimer's Disease since they provide insight into molecular processes in the brain\cite{marinescu2018tadpole}.
Our interpretation about the model not selecting the MRI features is that attention is distributed over 314 features which are indistinct compared to cognitive features. MRI features are valuable only when two scans taken over time between two visits to the hospital are compared to check the loss in volume.
However, in our case, we only consider scans at the baseline. 
For age prediction, the most relevant feature selected by our network was Volume of peripheral cortical grey matter, which is also supported by \cite{jiang2020predicting}.
\section{Discussion and Conclusion}
In this paper, we propose a novel GCN-based model that includes 1) generic interpretable attention module, 2) graph learning module, and 3) unique loss function. Our model learns the attention for input multi-modal features and uses the generated interpretation to train the model. IAM can be incorporated in any deep learning model capable of learning together with the task. 
The model's interpretation goes inline with the clinical findings and also the correlation with the ground truth. 
Further, our proposed GLM produces a latent population-level graph (heat maps added to supplementary) where we discovered 1) classes are clustered with a high number of edges between class samples within the class, 2) inter-class edges are important too as they help differentiate between the features from different classes.  

The end-to-end model leads to a significantly better classification performance than the state-of-the-art. Moreover, we showed the contribution of GLM, IAM, and the losses in our experiments. Our model is light and scalable in the case of multiple graph scenario. \cite{kazi2019self,kazi2019graph,coates2019mgmc}

\appendix

%
%
%
\bibliographystyle{splncs04}
\bibliography{mybibliography}
\end{document}


\section{Appendix}

\begin{figure}[t]
	\centering
	\includegraphics[width=0.7\textwidth]{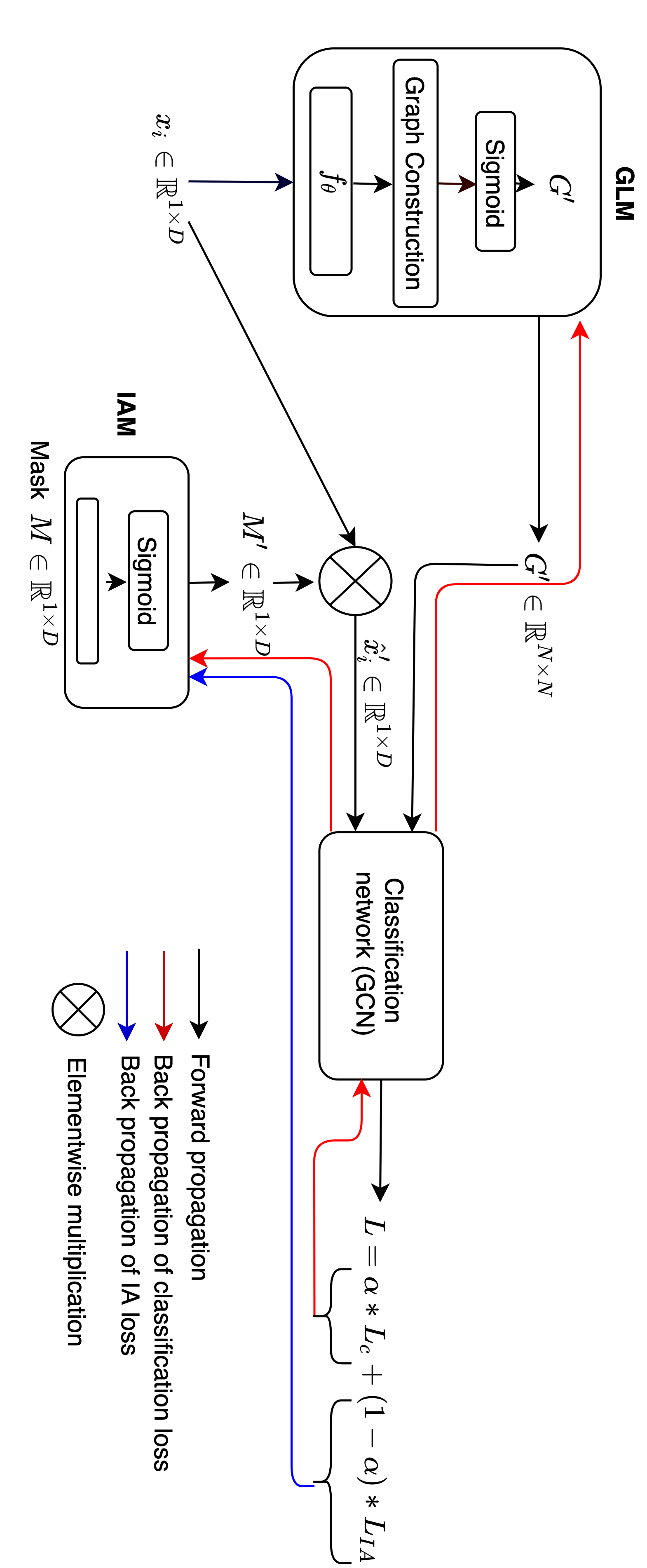}
	\caption{IA-GCN architecture. The model consists of three components: 1) Interpretable Attention Module(IAM), 2) Graph Learning Module(GLM), and 3) Classification Module. These are trained in an end-to-end training fashion. In backpropagation, two loss functions are playing roles which are demonstrated in blue and red arrows.}
	\label{fig.method}
\end{figure}

\textbf{List of feature chosen by IA-GCN for Tadpole for Alzehimer' prediction and their corresponding correlation coefficients and attention values reported by our model}
\begin{table}[htbp]
    \caption{IA-GCN on Tadpole dataset.}
      \centering
\begin{tabular}{ |c|c|c| } 
\hline
Feature&co-coefficient	&attention value on average\\
\hline
CDRSB&0.8166&.9202\\
CDRSB\_bl&0.8166&0.9193 \\
MMS&0.6697&0.4596 \\
MMSE\_bl&0.6702&0.4421 \\
others&0.15&0.093\\
\hline
\end{tabular}
\end{table}
\\
\textbf{List of feature chosen by IA-GCN for UKK for Gender prediction and their corresponding correlation coefficients and attention values reported by our model} 

\begin{table}[htbp]
    \caption{IA-GCN on UKBB dataset Gender prediction task.}
      \centering
\begin{tabular}{ |c|c|c| } 
\hline
Feature&co-coefficient	&attention value on average\\
\hline
Volumetric scaling from T1 head image to standard space&0.639&0.9561\\
Volume of white matter&0.575&0.947\\
Volume of brain, Grey+white matter&.532&0.8326\\
Volume of grey matter&0.427&0.6662\\
Volume of peripheral cortical grey matter&0.421&0.5149\\
\hline
\end{tabular}
\end{table}

\\
\textbf{List of feature chosen by IA-GCN for UKBB for age prediction and their corresponding correlation coefficients and attention values reported by our model}  

\begin{table}[htbp]
    \caption{IA-GCN on UKBB dataset age prediction task.}
      \centering
\begin{tabular}{ |c|c|c| } 
\hline
Feature&co-coefficient	&attention value on average\\
\hline
Volume of peripheral cortical grey matter (normalised for head size) & -0.539&0.5597\\
Mean MD in fornix on FA skeleton&0.418&0.7443\\
Mean L2 in fornix on FA skeleton&0.384&0.7244\\
Mean L3 in anterior corona radiata on FA skeleton (right)&0.225&0.6024\\
Mean L3 in anterior corona radiata on FA skeleton (left)&0.245&0.7449\\
Mean L3 in posterior corona radiata on FA skeleton (right)&0.193&0.5477\\
Mean L3 in posterior corona radiata on FA skeleton (left)&0.182&0.6731\\
Mean L3 in fornix cres+stria terminalis on FA skeleton (right)&0.259&0.732\\
Mean OD in fornix on FA skeleton&0.244&0.7504\\
\hline
\end{tabular}
\end{table}

\begin{figure}[t]
	\centering
	\includegraphics[width=1.0\textwidth]{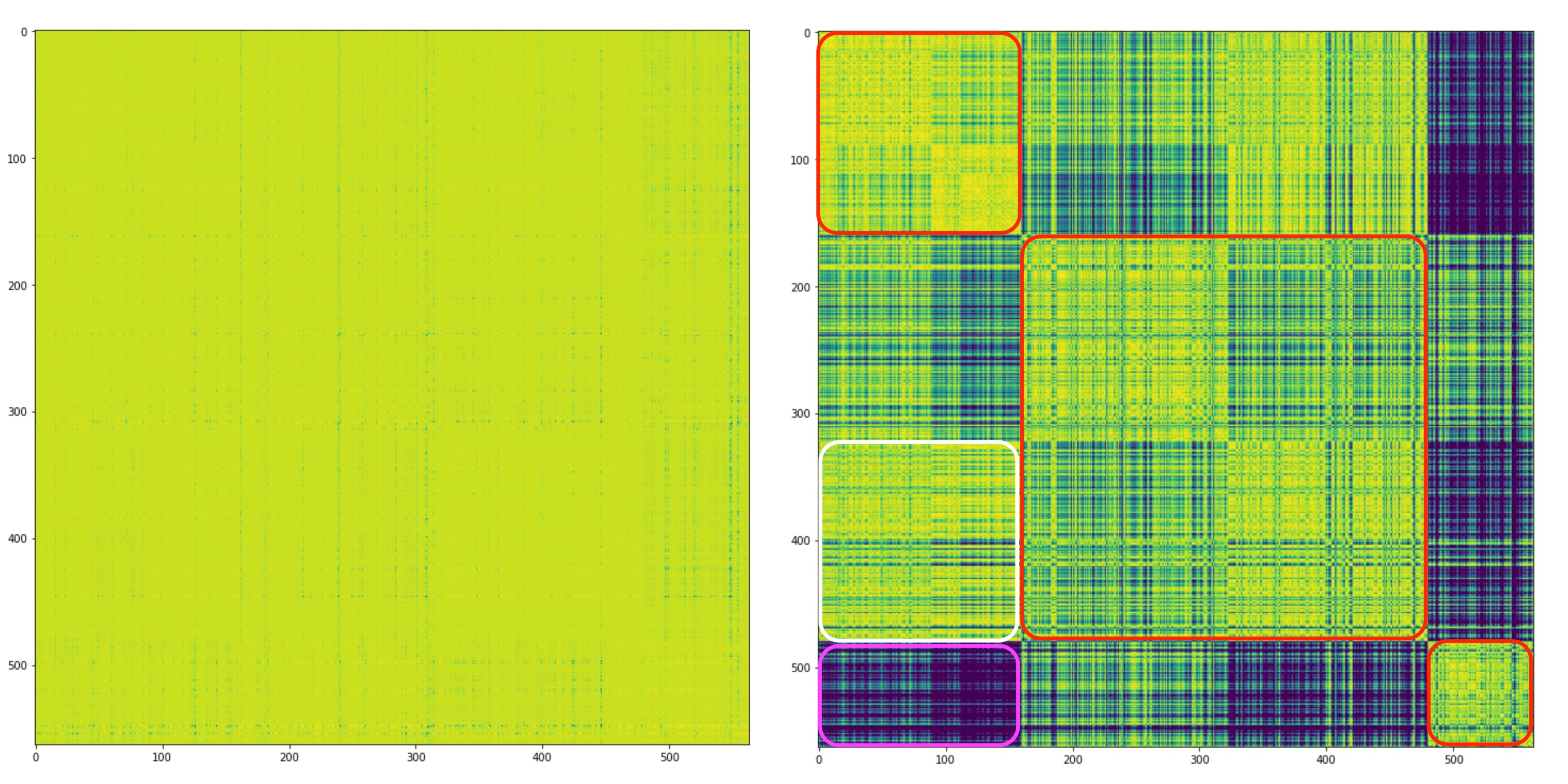}
	\caption{Heat maps result for the graph generated before training \textbf{(left)} and after training \textbf{(right)} by our GLM module.  It can be seen that in the beginning the graph is fully connected {(lefts)} which gain a semantic structure with inter- and intra- class connections. }
	\label{thresh}
\end{figure}

\textbf{Table 1} shows the behavior of the proposed IA-GCN for different  of contribution of classification loss $L_c$ ($\alpha$), $F_{MEL}$ ($\alpha_1$) and $F_{MSL}$ ($\alpha_2$)

\begin{table}[htbp]
    \caption{Performance of IA-GCN on Tadpole dataset in different settings.}
      \centering
\begin{tabular}{ |c|c|c|c|c|c| } 
\hline
$\alpha$& $\alpha_1$&$\alpha_2$ & Accuracy& Avg.4 & Avg.O\\
\hline
0 & 0.006 & 0.02 & 57.00±09.78 &$10^{-6}$& $10^{-6}$\\ 
0.2 & 0.006& 0.02 & 94.20±03.44 &00.12& 0.002 \\ 
0.4 & 0.006& 0.02 &   95.10±02.62&0.29& 0.001\\ 
\textbf{0.6} & \textbf{0.006}& \textbf{0.02}  & \textbf{96.10±02.49}&\textbf{0.74}& \textbf{0.09}\\
0.8 & 0.006& 0.02 &   95.80±02.31&0.78& 0.23\\
1.0 & 0.006& 0.02 &   95.40±02.32&0.82& 0.42\\ 
\hline

0.6& 0 & 0.02 & 95.60 ± 02.44 & 0.54 &0.13\\
0.6& 0.006 & 0 & 95.10 ± 03.69 &  0.86&0.21\\
\hline

\end{tabular}
\end{table}

\textbf{Table 2} shows the results of the GCN and DGM model for a) model trained conventionally, b) model trained with features selected by traditional dimensionality reduction method (ridge classifier), c) model trained on all features other than the ones selected by the proposed method and d) model trained only on the features selected by proposed method.  
\begin{table}[htbp]
\centering
\floatconts
  \label{tab:val alpha}%
  {\caption{Performance shown for classification task on Tadpole dataset. We show results for four baselines on GCN\cite{parisot2017spectral} and DGM\cite{cosmo2020latent} with different input feature setting.}}%
  {\begin{tabular}{|cc|c|c|c|c|} 
\hline
&&a&b&c&d\\
\hline
&Acc &77.4±02.41&81.00±06.40&74.50±3.44&82.4±04.14\\
\multirow{}{}{GCN}&AUC &79.79±04.75&74.70±04.32&72.11±08.24&83.89±09.06\\
&F1&74.7±05.32&78.4±06.77&65.23±08.46&78.73±07.60\\
\hline
&ACC&89.2±05.26&92.92±02.50&79.70±04.22&95.09±03.15\\
\multirow{}{}{DGM}&AUC&96.47±02.47&97.16±01.32&90.66±02.64&98.33±02.07\\
&F1&88.60±05.32&91.4±03.32&77.9±6.38&93.36±03.28\\
\hline
\end{tabular}}
\end{table}